\pdfoutput=1

\documentclass[11pt]{article}

\usepackage[]{acl}

\usepackage{times}
\usepackage{latexsym}

\usepackage[utf8]{inputenc}
\usepackage{booktabs}
\usepackage{graphicx}
\usepackage{subcaption}
\usepackage{wrapfig}
\usepackage{diagbox}
\usepackage{tikz}
\usepackage{makecell}
\usepackage{amsmath}
\usepackage{cleveref}
\crefformat{section}{\S#2#1#3} % see manual of cleveref, section 8.2.1
\crefformat{subsection}{\S#2#1#3}
\crefformat{subsubsection}{\S#2#1#3}
\usepackage{amssymb}
\usepackage{enumitem}
\usepackage{float}

\usepackage{color, colortbl}
\definecolor{Gray}{gray}{0.9}
\usepackage{multirow}

\usepackage[T1]{fontenc}

\usepackage[utf8]{inputenc}

\usepackage{microtype}

\title{Choose Your Lenses: Flaws in Gender Bias Evaluation}

\author{Hadas Orgad 
\hspace{10em}
  Yonatan Belinkov\thanks{~~Supported by the Viterbi Fellowship in the Center for Computer Engineering at the Technion.} \\
    \texttt{orgad.hadas@cs.technion.ac.il} \hspace{2em} 
  \texttt{belinkov@technion.ac.il} \\
  Technion -- Israel Institute of Technology}

\begin{document}
\maketitle
\begin{abstract}
Considerable efforts to measure and mitigate gender bias in recent years have led to the introduction of an abundance of tasks, datasets, and metrics used in this vein. 
In this position paper, we assess the current paradigm of gender bias evaluation and identify several flaws in it. First, we highlight the importance of extrinsic bias metrics that measure how a model's performance on some task is affected by gender, as opposed to intrinsic evaluations of model representations, which are less strongly connected to specific harms to people interacting with systems. We find that only a few extrinsic metrics are measured in most studies, although more can be measured. Second, we find that datasets and metrics are often coupled, and discuss how their coupling hinders the ability to obtain reliable conclusions, and how one may decouple them. We then investigate how the choice of the dataset and its composition, as well as the choice of the metric, affect bias measurement, finding significant variations across each of them. Finally, we propose several guidelines for more reliable gender bias evaluation.

\end{abstract}

\section{Introduction}

A large body of work has been devoted to measurement and mitigation of social biases in natural language processing (NLP), with a particular focus on gender bias  \cite{sun-etal-2019-mitigating, blodgett-etal-2020-language, garrido2021, stanczak2021survey}. These considerable efforts have been accompanied by various tasks, datasets, and metrics for evaluation and mitigation of gender bias in NLP models. In this position paper, we critically assess the predominant evaluation paradigm and identify several flaws in it. These flaws hinder progress in the field, since they make it difficult to ascertain whether progress has been actually made.

Gender bias metrics can be divided into two groups: \textit{extrinsic} metrics, such as performance difference across genders, measure gender bias with respect to a specific downstream task, while \textit{intrinsic} metrics, such as WEAT \cite{weat}, are based on the internal representations of the language model. We argue that measuring extrinsic metrics is crucial for building confidence in proposed metrics, defining the harms caused by biases found, and justifying the motivation for debiasing a model and using the suggested metrics as a measure of success. However, we find that many studies on gender bias only measure intrinsic metrics. As a result, it is difficult to determine what harm the presumably found bias may be causing. When it comes to gender bias mitigation efforts, improving intrinsic metrics may produce an illusion of greater success than reality, since their correlation to downstream tasks is questionable \cite{goldfarb-tarrant-etal-2021-intrinsic, intrinsic_extrinsic_contextual}. In the minority of cases where extrinsic metrics are reported, only few metrics are measured, although it is possible and sometimes crucial to measure more.

Additionally, gender bias measures are often applied as a \textit{dataset} coupled with a \textit{measurement technique} (a.k.a metric), but we show that they can be separated. A single gender bias metric can be measured using a wide range of datasets, and a single dataset can be applied to a wide variety of metrics. We then demonstrate how the choice of gender bias metric and the choice of dataset can each affect the resulting measures significantly. As an example, measuring the same metric on the same model with an imbalanced or a balanced dataset\footnote{Balanced with respect to the amount of examples for each gender, per task label.} may result in very different results. It is thus difficult to compare newly proposed metrics and debiasing methods with previous ones, hindering progress in the field.

To summarize, our contributions are:
\begin{itemize}
    \item We argue that extrinsic metrics are important for defining harms (\cref{sec:extrinsic}), but researchers do not use them enough even though they can (\cref{sec:diff_metrics}).
    \item We demonstrate the coupling of datasets with metrics and the feasibility of other combinations (\cref{sec:separation}).
    \item On observing that a specific metric can be measured on many possible datasets and vice-versa, we demonstrate how the choice and composition of a dataset (\cref{sec:datasets}), as well as the choice of bias metric to measure (\cref{sec:diff_metrics}), can strongly influence the measured results.
    \item We provide guidelines for researchers on how to correctly evaluate gender bias (\cref{sec:advise}).
\end{itemize}

\paragraph{Bias Statement}

 This paper examines metrics and datasets that are used to measure gender bias, and discusses several pitfalls in the current paradigm. As a result of the observations and proposed guidelines in this work, we hope that future results and conclusions will become clearer and more reliable.

The definition of gender bias in this paper is through the discussed metrics, as each metric reflects a different definition. Some of the examined metrics are measured on concrete downstream tasks (extrinsic metrics), while others are measured on internal model representations (intrinsic metrics). The definitions of intrinsic and extrinsic metrics do not align perfectly with the definitions of allocational and representational harms \cite{troublewithbias}. In the case of allocational harm, resources or opportunities are unfairly allocated due to bias. Representative harm, on the other hand, is when a certain group is negatively represented or ignored by the system. Extrinsic metrics can be used to quantify both allocational and representational harms, while intrinsic metrics can only quantify representational harms, in some cases.

There are also other important pitfalls that are not discussed in this paper, like the focus on high-resource languages such as English and the binary treatment of gender \cite{sun-etal-2019-mitigating, stanczak2021survey, dev-etal-2021-harms}. Inclusive research of non-binary genders would require a new set of methods, which could benefit from the observations in this work.

\section{The Importance of Extrinsic Metrics in Defining Harms}
\label{sec:extrinsic}

In this paper, we divide metrics for gender bias to three groups:

\begin{itemize}[itemsep=3pt,topsep=3pt,parsep=3pt]
    \item \textbf{Extrinsic performance:} measures how a model's performance is affected by gender, and is calculated with respect to particular gold labels. For example, the True Positive Rate (TPR) gap between female and male examples.
    \item \textbf{Extrinsic prediction:} measures model's predictions, such as the output probabilities, but the bias is not calculated with respect to some gold labels. Instead, the bias is measured by the effect of gender or stereotypes on model predictions. For example, the probability gap can be measured on a language model queried on two sentences, one pro-stereotypical (``he is an engineer'') and another anti-stereotypical (``she is an engineer'').
    \item \textbf{Intrinsic:} measures bias in internal model representations, and is not directly related to any downstream task. For example, WEAT.
\end{itemize}

It is crucial to define how measured bias harms those interacting with the biased systems \cite{barocas2017problem, troublewithbias, blodgett-etal-2020-language, bommasani2021opportunities}. Extrinsic metrics are important for motivating bias mitigation and for accurately defining ``why the system behaviors that are described as `bias' are harmful, in what ways, and to whom'' \cite{blodgett-etal-2020-language}, since they clearly demonstrate the performance disparity between protected groups.

For example, in a theoretical CV-filtering system, one can measure the TPR gap between female and male candidates. A gap in TPR favoring men means that, given a set of valid candidates, the system picks valid male candidates more often than valid female candidates. The impact of this gap is clear: Qualified women are overlooked because of bias. In contrast, consider an intrinsic metric such as  WEAT \cite{weat}, which is derived from the proximity (in vector space) of words like ``career'' or ``family'' to ``male'' or ``female'' names. If one finds that male names relate more to career and female names relate more to family, the consequences are unclear. In fact, \citet{goldfarb-tarrant-etal-2021-intrinsic} found that WEAT does not correlate with other extrinsic metrics. However, many studies report only intrinsic metrics (a third of the papers we reviewed, \cref{sec:diff_metrics}).

\section{Coupling of Datasets and Metrics}

\label{sec:separation}
In this section, we discuss how datasets and metrics for gender bias evaluation are typically coupled, how they may be decoupled, and why this is important. 
We begin with a representative test case, followed by a discussion of the general phenomenon.

\subsection{Case study: Winobias}

\begin{figure}[t]
  \includegraphics[width=\columnwidth]{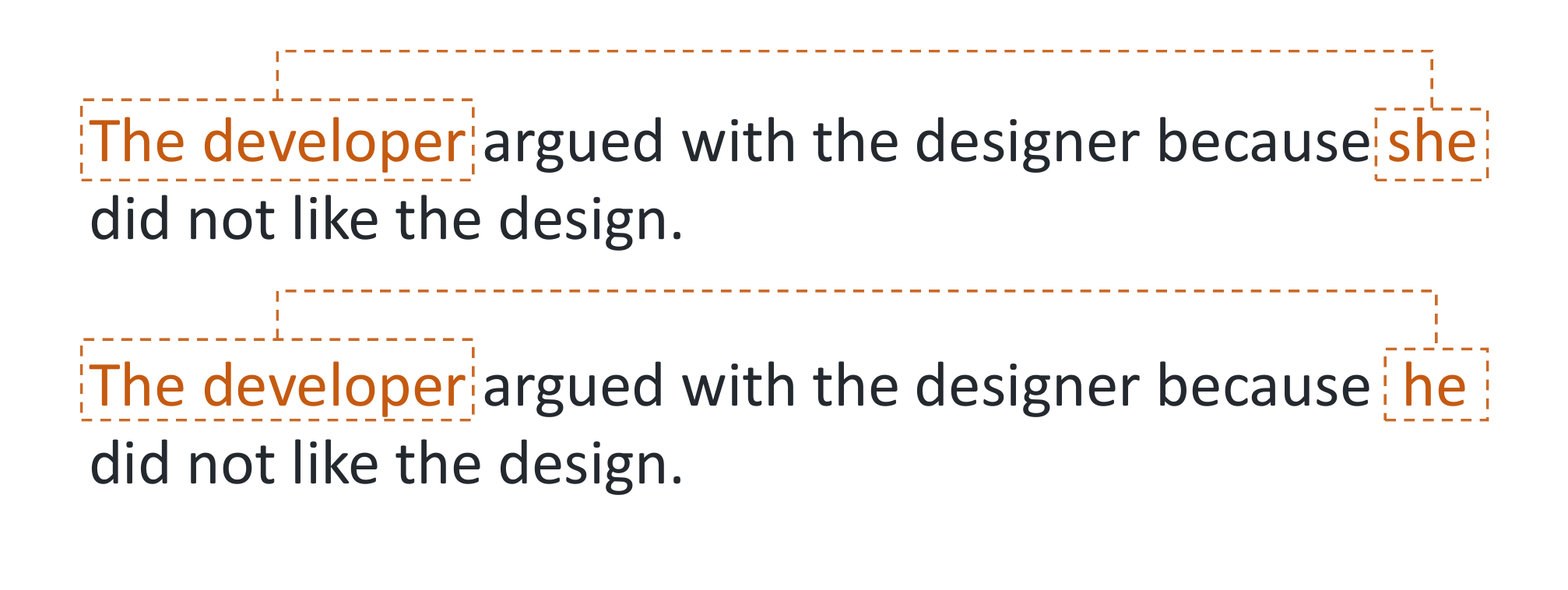}
  \caption{Coreference resolution example from Winobias: a pair of anti-stereotypical (top) and pro-stereotypical examples (bottom). Developers are stereotyped to be males.}
  \label{fig:winobias}
  \vspace{-10pt}
\end{figure}

Coreference resolution aims to find all textual expressions that refer to the same real-world entities. A popular dataset for evaluating gender bias in coreference resolution systems is Winobias \cite{winobias}. It consists of Winograd schema \cite{levesque2012winograd} instances: two sentences that differ only by one or two words, but contain ambiguities that are resolved differently in the two sentences based on world knowledge and reasoning. Winobias sentences consist of an anti- and a pro- stereotypical sentence, as shown in Figure \ref{fig:winobias}. Coreference systems should be able to resolve both sentences correctly, but most perform poorly on the anti-stereotypical ones \cite{winobias, zhao-etal-2019-gender, de-vassimon-manela-etal-2021-stereotype, our_paper}.

Winobias was originally proposed as an extrinsic evaluation dataset, with a reported metric of anti- and pro- stereotypical performance disparity. However, other metrics can also be measured, both intrinsic and extrinsic, as shown in several studies \cite{zhao-etal-2019-gender, nangia-etal-2020-crows,our_paper}. For example, one can measure how many stereotypical choices the model preferred over anti-stereotypical choices (an extrinsic performance measure), as done on Winogender \cite{winogender}, a similar dataset. Winobias sentences can also be used to evaluate language models (LMs), by evaluating if an LM gives higher probabilities to pro-stereotypical sentences \cite{nangia-etal-2020-crows} (an extrinsic prediction measure). Winobias can also be used for intrinsic metrics, for example as a template for SEAT \cite{seat} and CEAT \cite{ceat} (contextual extensions of WEAT). Each of these metrics reveals a different facet of gender bias in a model. An explicit measure of how many pro-stereotypical choices were preferred over anti-stereotypical choices has a different meaning than measuring a performance metric gap between two different genders. Additionally, measuring an intrinsic metric on Winobias may be help tie the results to the model's behavior on the same dataset in the downstream coreference resolution task.

\subsection{Many possible combinations for datasets and metrics}
\begin{table*}[h]
\Large
\centering
\resizebox{\textwidth}{!}{
\begin{tabular}{lllllllllllllll}
\toprule
                & \multicolumn{3}{c}{\textbf{Extrinsic Performance}} & \multicolumn{3}{c}{\textbf{Extrinsic Predictions}} & \multicolumn{6}{c}{\textbf{Intrinsic}} \\
\cmidrule(r){2-4} \cmidrule(r){5-8} \cmidrule(r){9-15}
\backslashbox{\textbf{Dataset}}{\textbf{Metric}} &  Gap & Gap & Gap & \% or \# of &  \% or \# Model & Pred  & LM Prediction & SEAT & CEAT & Probe & Cluster & Nearest & Cos & PCA \\
& (Label) & (Stereo) & (Gender) & Answer Changed & Prefers Stereotype & Gap & On Target words & & & & & Neighbors & \\
\midrule
\rowcolor{Gray}
Winogender \cite{winogender} & \checkmark & \textcircled{$\checkmark$} & \checkmark & \checkmark & \checkmark & \checkmark & \checkmark &  & \checkmark & \checkmark & \checkmark & \checkmark & \checkmark & \checkmark \\
Winobias \cite{winobias} & \checkmark & \textcircled{$\checkmark$} & \checkmark & \checkmark & \checkmark & \checkmark & \checkmark & & \checkmark & \checkmark & \checkmark & \checkmark & \checkmark & \checkmark \\
\rowcolor{Gray}
Gap \cite{gap} & & & \textcircled{$\checkmark$} & \checkmark (aug) & & & & & & & & & & \\
Crow-S \cite{crows} & & & & & \textcircled{$\checkmark$} & & \checkmark & & \checkmark & \checkmark & \checkmark & \checkmark & \checkmark & \checkmark \\
\rowcolor{Gray}
StereoSet \cite{stereoset} & & & & & \textcircled{$\checkmark$} & & \checkmark & & \checkmark & \checkmark & \checkmark & \checkmark & \checkmark & \checkmark \\
Bias in Bios \cite{biasinbios} & \textcircled{$\checkmark$} & \checkmark & \checkmark & \checkmark (aug) & \checkmark (aug) & \checkmark (aug) & \checkmark & & \checkmark & \checkmark & \checkmark & \checkmark & \checkmark & \checkmark\\
\rowcolor{Gray}
EEC \cite{eec} & & & & & & \textcircled{$\checkmark$} & \checkmark & & \checkmark & \checkmark & \checkmark & \checkmark & & \\
STS-B for genders \cite{disco} & & & & & \checkmark & \textcircled{$\checkmark$} & \checkmark & \checkmark & \checkmark & \checkmark & \checkmark & \checkmark & \checkmark & \checkmark \\
\rowcolor{Gray}
\citet{gendered_nli} (NLI) & & & & \checkmark & \checkmark & \checkmark & \checkmark & & \checkmark & \checkmark & \checkmark & \checkmark & \checkmark & \checkmark \\
PTB, WikiText, CNN/DailyMail & & & & & & & \textcircled{$\checkmark$} & & \checkmark\\
\cite{bordia-bowman-2019-identifying} \\
\rowcolor{Gray}
BOLD \cite{bold} & & & & & & & \textcircled{$\checkmark$} & & & & & & & \\
Templates from \citet{seat} & & & & & & & $\checkmark$ &  \textcircled{$\checkmark$} & \checkmark & \checkmark & \checkmark & \checkmark & \checkmark & \checkmark \\
\rowcolor{Gray}
Templates from \citet{kurita-etal-2019-measuring} & & & & & & & \textcircled{$\checkmark$} &  \textcircled{$\checkmark$} & \checkmark & \checkmark & \checkmark & \checkmark & \checkmark & \checkmark \\
DisCo templates \cite{disco} & & & & & \checkmark & & \textcircled{$\checkmark$} & & \checkmark & \checkmark & \checkmark & \checkmark & \checkmark & \checkmark  \\
\rowcolor{Gray}
BEC-Pro templates \cite{bartl-etal-2020-unmasking} & & & & & \checkmark & & \textcircled{$\checkmark$} & & \checkmark & \checkmark & \checkmark & \checkmark & \checkmark & \checkmark \\
English-German news corpus  & & & & & & & \checkmark & & \checkmark & \textcircled{$\checkmark$} & \textcircled{$\checkmark$} & \textcircled{$\checkmark$} & \textcircled{$\checkmark$} & \textcircled{$\checkmark$} \\
\cite{basta2021extensive} \\ 
\rowcolor{Gray}
Reddit (\citealt{ceat}, & & & & & & & \checkmark & & \textcircled{$\checkmark$} & \checkmark & \checkmark & \checkmark & \checkmark & \checkmark \\
\rowcolor{Gray}
\citealt{voigt-etal-2018-rtgender}) & & & & & & & & & & & & & & \\
MAP \cite{cao-daume-iii-2021-toward} & & & \textcircled{$\checkmark$} & \checkmark & & & & & \checkmark & \checkmark & \checkmark & \checkmark & \checkmark & \checkmark \\
\rowcolor{Gray}
GICoref \cite{cao-daume-iii-2021-toward} & & & \textcircled{$\checkmark$} & & & & & & \checkmark & \checkmark & \checkmark & \checkmark & \checkmark & \checkmark \\
\bottomrule
\end{tabular}
}
    \caption{Combinations of gender bias datasets and metrics in the literature. $\checkmark$ marks a feasible combination of a metric and a dataset. \textcircled{$\checkmark$} marks the original metrics used on the dataset, and \checkmark (aug) marks metrics that can be measured after augmenting the dataset such that every example is matched with a counterexample of another gender. Extrinsic performance metrics depend on gold labels while extrinsic prediction metrics do not. A full description of the metrics is given in Appendix \ref{app:metrics}.}
    \label{metrics}
    \vspace{-10pt}
\end{table*}
Winobias is one example out of many. In fact, benchmarks for gender bias evaluation are typically proposed as a package of two components:
\begin{enumerate}[itemsep=3pt,topsep=3pt,parsep=3pt]
    \item \textbf{A dataset} on which the benchmark task is performed.
    \item \textbf{A metric}, which is the particular method used to calculate bias of a model on the dataset.
\end{enumerate}

Usually, these benchmarks are considered as a bundle; however, they can often be decoupled, mixed, and matched, as discussed in the Winobias test case above. The work by \citet{delobelle2021measuring} is an exception, in that they gathered a set of templates from diverse studies and tested them using the same metric. 

In Table \ref{metrics}, we present possible combinations of datasets (rows) and metrics (columns) from the gender bias literature. The metrics are partitioned according to the three classes of metrics defined in Section \ref{sec:extrinsic}. We present only metrics valid for assessing bias in contextualized LMs (rather than static word embeddings), since they are the common practice nowadays. The table does not claim to be exhaustive, but rather illustrates how metrics and datasets can be repurposed in many different ways. The metrics are described in appendix \ref{app:metrics}, but the categories are very general and even a single column like ``Gap (Label)'' represents a wide variety of metrics that can be measured.

Table \ref{metrics} shows that many metrics are compatible across many datasets (many  $\checkmark$'s in the same column), and that datasets can be used to measure a variety of metrics other than those typically measured (many  $\checkmark$'s in the same row). Some datasets, such as Bias in Bios \cite{biasinbios}, have numerous metrics compatible, while others have fewer, but still multiple, compatible metrics. Bias in Bios has many compatible metrics since it has information that can be used to calculate them: in addition to gold labels, it also has gender labels and clear stereotype definitions derived from the labels which are professions. Text corpora and template data, which do not address a specific task (bottom seven rows), are mostly compatible with intrinsic metrics. The compatibility of intrinsic metrics with many datasets may explain why papers report intrinsic metrics more often (\cref{sec:diff_metrics}). Additionally, Table \ref{metrics} indicates that not many datasets can be used to measure extrinsic metrics, particularly extrinsic performance metrics that require gold labels. On the other hand, measuring LM prediction on target words, which we consider as extrinsic, can be done on many datasets. This is useful for analyzing bias when dealing with LMs. It can be done by computing bias metrics from the LM output predictions, such as the mean probability gap when predicting the word ``he'' versus ``she'' in specific contexts. Also, some templates are valid for measuring extrinsic prediction metrics, especially stereotype-related metrics, as they were developed with explicit stereotypes in mind (such as profession-related stereotypes).

Based on Table \ref{metrics}, it is clear that there are many possible ways to measure gender bias in the literature, but they all fall under the vague category of ``gender bias''. Each of the possible combinations gives a different definition, or interpretation, for gender bias. The large number of different metrics makes it difficult or even impossible to compare different studies, including proposed gender bias mitigation methods. This raises questions about the validity of results derived from specific combinations of measurements. In the next two sections, we demonstrate how the choice of datasets and metrics can affect the bias measurement.

\section{Effect of Dataset on Measured Results}
\label{sec:datasets}

The choice of data to measure bias has an impact on the calculated bias. Many researchers used sentence templates that are ``semantically bleached'' (e.g., ``This is <word>.'', ``<person> studied <profession> at college.'') to adjust metrics developed for static word embeddings to contextualized representations  \cite{may-etal-2019-measuring, kurita-etal-2019-measuring, webster2020measuring, bartl-etal-2020-unmasking}. \citet{delobelle2021measuring} found that the choice of templates significantly affected the results, with little correlation between different templates. Additionally, \citet{may-etal-2019-measuring} reported that templates are not as semantically bleached as expected. 

Another common feature of bias metrics is the use of hand-curated word lexicons by almost every bias metric in the literature. \citet{antoniak-mimno-2021-bad} reported that the lexicon choice can greatly affect bias measurement, leading to differing conclusions between different lexicons.

\vspace{-3pt}
\subsection{Case study: balancing the test data}

Another important variable in gender bias evaluation, often overlooked in the literature, is the composition of the test dataset. Here, we demonstrate this by comparing metrics on different test sets, which come from the same dataset but have a different balance of examples. Bias in Bios \cite{biasinbios} involves predicting an occupation from a biography text. These occupations are not balanced across genders, so for example over 90\% of the nurses in the dataset identify as females.

Our case study extends the experiments done by \citet{our_paper}. In their work, they tested a RoBERTa-based \cite{roberta} classifier fine-tuned on Bias in Bios. The model was trained and evaluated on a training/test split of the dataset using numerous extrinsic bias metrics. Here we train the same model on the same training set, but evaluate it on three types of test sets: the original test set alongside two balanced versions of it, which have equal numbers of females and males in every profession, by either subsampling or oversampling. \footnote{Subsampling is the process of removing examples from the dataset such that the resulting dataset contains the same number of male and female examples for each label. Oversampling achieves this by repeating examples.} We follow \citeauthor{our_paper} and report nine different metrics on this task, measuring either some notion of performance gap or a statistical metric from the fairness literature. For details on the metrics measured in this experiments, see Appendix \ref{app:impl}.

\begin{table}

\centering
\resizebox{\columnwidth}{!}{
\begin{tabular}{llrrr}
\toprule
\multirow{2}{*}{Metric} & \multicolumn{3}{c}{Testing balancing} \\
\cmidrule{2-4}
 &  Original &  Oversampled &  Subsampled \\
\midrule
TPR (p) &      0.78 &         0.75 &        0.75 \\
TPR (s) &      2.35 &         2.41 &        2.38 \\
FPR (p) &      0.61 &         0.59 &        0.57 \\
FPR (s) &      0.08 &         0.08 &        0.08 \\
Precision (p) &      0.71 &         -0.80 &        -0.89* \\
Precision (s) &      1.87 &         2.59* &        3.69* \\
\arrayrulecolor{gray}\hline
Separation (s) &      2.27 &         0.23* &        0.35* \\
Sufficiency (s) &      1.94 &         0.74* &        9.15* \\
Independence (s) &      0.14 &         0.01* &        0.01* \\
\bottomrule
\end{tabular}
}
\caption{Metrics measured on Bias in Bios, separated to performance gap metrics (above the line) and statistical fairness metrics (below the line). Metrics are measured on the original test split, and on a subsampled and oversampled version of it. * marks statistically significant difference in a metric compared to the baseline (Original), using Pitman's permutation test ($p < 0.05$).}
\vspace{-10pt}
\label{tbl:test_balancing}
\end{table}

\begin{figure}[t]
  \includegraphics[width=.8\columnwidth]{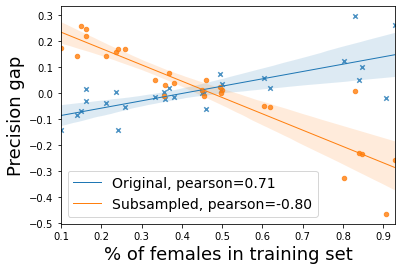}
  \caption{Percentage of females in the training set versus the resulting precision gap, per each profession. The trend is opposite on different test sets.}
  \label{fig:tpr_bios_corr_compare_before}
  \vspace{-15pt}
\end{figure}

As the results in Table \ref{tbl:test_balancing} show, although many of the gap metrics (top block) are unaffected by the balancing of the test dataset, the absolute sum of precision gaps is significantly different when tested on different datasets. Moreover, the Pearson correlation for precision is negative when tested on a balanced dataset (either subsampled or oversampled), which may lead to different conclusions. The Pearson correlation is computed between the performance gaps per label (profession), and the percentage of females in the training set for that label, without balancing (the original distribution can be found in Appendix \ref{app:bias-in-bios-stats}). Strong correlations (positive or negative) indicate more bias was learned from the training set, and the direction of bias is indicated by the sign. This correlation is illustrated in Figure \ref{fig:tpr_bios_corr_compare_before} for the original (imbalanced) test set and a balanced (subsampled) test set. Appendix \ref{app:precision-analysis} discusses why this occurs.

%  of professions per gender

The statistical fairness metrics (bottom block in Table \ref{tbl:test_balancing}) show a significant difference in the measured bias across different test set balancing. Oversampling shows less bias than when measured on the original test set, while subsampling yields mixed results -- it decreases one metric while increasing another.

\paragraph{What is the ``correct'' test set?} Since metrics are defined over the entire dataset, they are sensitive to its composition. For measuring bias in a model, the dataset used should be as unbiased as possible, thus balanced datasets are preferable.

\begin{figure*}[t!]
\begin{subfigure}[t]{\columnwidth}
  \includegraphics[width=\columnwidth]{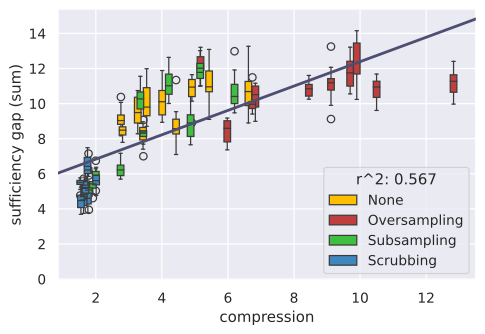}
  \caption{Intrinsic metric was measured on the test set of occupation prediction, figure reproduced from \citet{our_paper}.}
  \label{fig:bios_compression_sufficiency}
\end{subfigure}
\hspace{1em}
\begin{subfigure}[t]{\columnwidth}
  \includegraphics[width=\columnwidth]{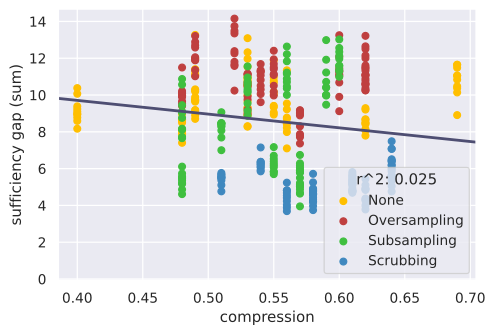}
  \caption{Intrinsic metric was measured on Winobias \cite{winobias}.}
  \label{fig:winobias_compression_sufficiency}
\end{subfigure}
\caption{Correlation between an intrinsic metric (compression) and an extrinsic metric (sufficiency gap sums), for various models trained on occupation prediction task.``None'' was trained on the original dataset, ``Oversampling'' was trained on an oversampled dataset, ``Subsampling'' was trained on a subsampled dataset and ``Scrubbing'' was trained on a scrubbed dataset (explicit gender words like ``he'' and ``she'' were removed).}
\label{fig:sufficiency_gap}
\vspace{-10pt}
\end{figure*}

If we were only concerned with measuring one of the reduced metrics on a non-balanced test set, we could misrepresent the fairness of the model. Indeed, it is common practice to measure only a small portion of metrics out of all those that can be measured---as we show in section \ref{sec:diff_metrics}---which makes us vulnerable to misinterpretations.

\subsection{Case study: measuring intrinsic bias on two different datasets}

It is critical to consider the impact of the data used when measuring intrinsic bias metrics on a language model. Previous work \cite{goldfarb-tarrant-etal-2021-intrinsic, intrinsic_extrinsic_contextual, our_paper} inspected the correlations between extrinsic and intrinsic gender bias metrics. Some did not find correlations, while others did in some cases. However, correlations do not solely depend on the model used for bias measurement, but also on the dataset used to measure the intrinsic metric.

Our experiment analyzes the behavior of the same metric on different datasets. We again follow \citet{our_paper}, who probed for the amount of gender information extractable from the model's internal representations. This is quantified by \emph{compression} \cite{voita-titov-2020-information}, where a higher compression indicates greater bias extractability. \citeauthor{our_paper} found that this metric correlates strongly with various extrinsic metrics. An example of this correlation is shown in Figure \ref{fig:bios_compression_sufficiency} on the Bias in Bios task with models debiased with various strategies. The correlation is high ($r^2=0.567$).

In their experiment, the intrinsic metric was measured on the same dataset as the extrinsic one. We repeat the correlation tests, but this time measure the intrinsic metric on a different dataset, the Winobias dataset. The results (Figure \ref{fig:winobias_compression_sufficiency}) clearly show that there is no correlation between extrinsic and intrinsic metrics in this case ($r^2=0.025$).

Hence, we conclude that the dataset used to measure intrinsic bias impacts the results significantly. To reliably reflect the biases that the model has acquired, it should be closely related to the task that the model was trained on. In our experiment, when intrinsic and extrinsic metrics were not measured on the same dataset, no correlation was detected. This is the case for all metrics on this task from \citet{our_paper}; see Appendix \ref{tbl:other_data_correlations}. As discussed in \cref{sec:separation}, the same intrinsic metrics can be evaluated across a variety of datasets. Even so, some intrinsic metrics were originally defined to be measured on different datasets than the task dataset, such as those defined on templates (Table \ref{metrics}).

\vspace{-3pt}
\section{Different Metrics Cover Different Aspects of Bias}
\label{sec:diff_metrics}

\begin{figure}[t]
  \centering
  \includegraphics[width=.8\columnwidth]{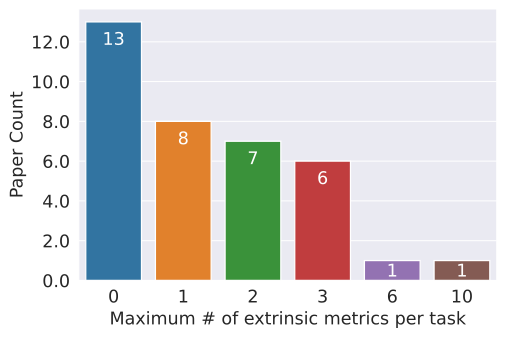}
  \caption{The number of extrinsic metrics measured in the papers we reviewed.}
  \label{fig:extrinsic_count}
\vspace{-15pt}
\end{figure}

In this section, we explore how the choice of bias metrics influences results. Although extrinsic bias metrics are useful in defining harms caused by a gender-biased system, we find that most studies on gender bias use only intrinsic metrics to support their claims. We surveyed a representative list of papers presenting bias mitigation techniques that appeared in the survey by \citet{stanczak2021survey}, as well as recent papers from the past year. In total, we examined 36 papers. Many papers do not measure extrinsic metrics. Even when downstream tasks are measured, only a very small subset of metrics (three or less) is typically measured, as shown in Figure \ref{fig:extrinsic_count}. Furthermore, in these studies, typically no explanation is provided for choosing a particular metric.

The exceptions are \citet{de-vassimon-manela-etal-2021-stereotype} and \citet{our_paper}, who measured six and nine or ten metrics on downstream tasks, respectively. \citeauthor{our_paper} showed that different extrinsic metrics behave differently under various debiasing methods. Additionally, in \cref{sec:datasets} we saw that subsampling the test set increased one bias metric and decreased others, which would not have been evident had we only measured a small number of metrics. Measuring multiple metrics is also important for evaluating debiasing. When \citeauthor{kaneko-bollegala-2021-debiasing} compared their proposed debiasing method to that of \citet{gendered_nli}, the new method outperformed the old one on two of the three metrics.

As the examples above illustrate, different extrinsic metrics are not necessarily consistent with one another. Furthermore, it is possible to measure more extrinsic metrics, although it is rarely done. When it is not feasible to measure multiple metrics, one should at least justify why a particular metric was chosen. In a CV-filtering system, for example, one might be more forgiving of FPR gaps than of TPR gaps, as the latter leaves out valid candidates for the job in one gender more than the other. However, more extrinsic metrics are likely to provide a more reliable picture of a model's bias.

\section{Conclusion and Proposed Guidelines}
\label{sec:advise}

The issues described in this paper concern the instabilities and vagueness of gender bias metrics in NLP. Since bias measurements are integral to bias research, this instability limits progress. We now provide several guidelines for improving the reliability of gender bias research in NLP.

\paragraph{Focus on downstream tasks and extrinsic metrics.} Extrinsic metrics are helpful in motivating bias mitigation (\cref{sec:extrinsic}). However, few datasets can be used to quantify extrinsic metrics, especially extrinsic performance metrics, which require gold labels (\cref{sec:separation}). More effort should be devoted to collecting datasets with extrinsic bias assessments, from more diverse domains and downstream tasks.

\paragraph{Stabilize the metric or the dataset.} Both the metrics and the datasets could have significant effects over the results: The same dataset can be used to measure many metrics and yield different conclusions (\cref{sec:datasets}), and the same metric can be measured on different datasets that lead to different results and different conclusions (\cref{sec:diff_metrics}). If one wishes to measure gender bias in an NLP system, it is better to hold one of these variables fixed: for example, to focus on a single metric and measure it on a set of datasets. Of course, this can be repeated for other metrics as well. This will produce much richer, more consistent, and more convincing results.

\paragraph{Neutralize dataset noise.} As a result of altering a dataset's composition, we observed very different results (\cref{sec:datasets}). This is caused by the way various fairness metrics are defined and computed on the entire dataset. To ensure a more reliable evaluation, we recommend normalizing a dataset when using it for evaluation. In the case of occupation prediction, normalization can be obtained by balancing the test set. In other cases it could be by anonymizing the test set, removing harmful words, etc., depending on the specific scenario.

\paragraph{Motivate the choice of specific metrics, or measure many.} Most work measures only a few metrics  (\cref{sec:diff_metrics}). A comprehensive experiment, such as to prove the efficacy of a new debiasing method, is more reliable if many metrics are measured. In some situations, a particular metric may be of interest; in this case one should carefully justify the choice of metric and the harm that is caused when the metric indicates bias. The motivation for debiasing this metric then follows naturally.

\paragraph{Define the motivation for debiasing through bias metrics.}
\citet{blodgett-etal-2020-language} found that papers’ motivations are ``often vague, inconsistent, and lacking in normative reasoning''. We propose to describe the motivations through the gender bias metrics chosen for the study: define what is the harm measured by a specific metric, what is the behavior of a desired versus a biased system, and how the metric measures it. This is where extrinsic metrics will be particularly useful.

\medskip 
We believe that following these guidelines will enhance clarity and comparability of results, contributing to the advancement of the field.

\section*{Acknowledgements}
\vspace{-5pt}
This research was supported by the ISRAEL SCIENCE FOUNDATION (grant No.\ 448/20) and by an Azrieli Foundation Early Career Faculty Fellowship. We also thank the anonymous reviewers for their insightful comments and suggestions, and the members of the Technion CS NLP group for their valuable feedback.

\bibliography{anthology,custom}
\bibliographystyle{acl_natbib}

\appendix

\section{List of gender bias metrics, as presented in Table \ref{metrics}}
\label{app:metrics}

Many of the items in this list do not aim to describe a specific metric, but rather describe a family of metrics with similar characteristics and requirements.

\subsection{Extrinsic Performance}
This class of extrinsic metrics measures how a model's performance is affected by gender. This is computed with respect to particular gold labels and there is a clear defintion of harm derived from the specific performance metric measured, for instance F1, True Positive Rate (TPR), False Positive Rate (FPR), BLEU score for translation tasks, etc.
\begin{enumerate}
    \label{app:gap_label}
    \item \textbf{Gap (Label):} Measures the difference in some performance metric between Female and Male examples, in a specific class. The performance gap can be computed as the difference or the quotient between two performance metrics on two protected group. For example, in Bias in Bios \cite{biasinbios} one can measure the TPR gap between female teachers and male teachers. The gaps per class can be summed, or the correlation with the percentage of women in the particular class can be measured.
    \item \textbf{Gap (Stereo):} Measures the difference in some performance metric between pro-stereotypical (and/or non-stereotypical) and anti-stereotypical (and/or non-stereotypical) instances. A biased model will have better performance on pro-stereotypical instances. This can be measured across the whole dataset or per gender / class.
    \item \textbf{Gap (Gender):} Measure the difference in some performance metric between male examples and female examples, across the entire dataset. In cases of non-binary gender datasets \cite{cao-daume-iii-2021-toward}, the gap can be calculated to measure the difference between text that is trans-inclusive versus text that is trans-exclusive. Another option is to measure the difference in performance before and after removing various aspects of gender from the text.
\end{enumerate}

\subsection{Extrinsic Prediction}
This class is also extrinsic as it measures model predictions, but the bias is not computed with respect to some gold labels. Instead, the bias is measured by the effect of gender on the predictions of the model.
\begin{enumerate}
    \item \textbf{\% or \# of answer changes:} The number or percentage that the prediction changed when the gender of the example changed. To measure this, each example should have a counterpart example of the opposite gender. This difference can be measured with respect to the number of females or males in the specific label, for instance with relation to occupation statistics.
    \item \textbf{\% or \# that model prefers stereotype:} Quantifies how much the model tends to go for the stereotypical option, for instance predicting that a ``she'' pronoun refers to a nurse in a coreference resolution task. This can also be measured as a correlation with the number of females or males in the label, which can be thought of as the ``strength'' of the stereotype.
    
    \item \textbf{Pred gap:} The raw probabilities or some function of them are measured, and the gap is measured as the prediction gap between male and female predictions. This can be measured across the whole dataset or per label at other cases.
    
    \item \textbf{LM prediction on target words}: This metric relates to the specific predictions of a pre-trained LM, such as a masked LM. The prediction of the LM is calculated for a specific text or on a specific target word of interest. These probabilities are then used to measure the bias of the model.  We did not include this metric category in the ``Pred gap'' category  because it can be measured on a much larger number of datasets. For example, for the masked sentence: ``The programmer said that <mask> would finish the work tomorrow'', we might measure the relation between $p(<mask>=he|sentence)$ and $p(<mask>=she|sentence)$. Although somewhat similar in idea to the previously described metric ``pred gap'', it is presented as a separate metric since it can be computed on a wider range of datasets. The strategy for calculating a number quantifying bias from the raw probabilities varies in different papers. For example, \citet{kurita-etal-2019-measuring, crows, bordia-bowman-2019-identifying, stereoset} all use different formulations.
\end{enumerate}

\subsection{Intrinsic}
This class measures bias on the hidden model representations, and is not directly related to any downstream task.
\begin{enumerate}

    \item \textbf{WEAT:} The Word Embedding Association Test \cite{weat} was proposed as a way to quantify bias in static word embeddings. While we consider only bias metrics that can be applied in contextualized settings, we describe WEAT here as it is popular and has been adapted to contextualized settings. To compute WEAT, one defines a set of target words $X, Y$ (e.g., programmer, engineer, scientist, etc., and nurse, teacher, librarian, etc.)\ and two sets of attribute words $A, B$ (e.g., man, male, etc. and woman, female, etc.). The null hypothesis is that the two sets of target words are not different when it comes to their similarity to the two sets of attribute words. We test the null hypothesis using a permutation test on the word embeddings, and the resulting effect size is used to quantify how different the two sets are.

    \item \textbf{SEAT:} the Sentence Encoder Association Test \cite{seat} was proposed as a contextual version of the popular metric WEAT. As WEAT was computed on static word embedding, in SEAT they proposed using ``semantically-bleached'' templates such as ``This is [target]'', where the target word of interest is planted in the template, to get its word embedding in contextual language models. Thus, we only consider ``semantically-bleached'' templates to be appropriate as a dataset for SEAT.
    
    \item \textbf{CEAT:} Contextualized Embedding Association Test \cite{ceat} was proposed as another contextual alternative to WEAT. Here, instead of using templates to get the word of interest, for each word a large number of embeddings is collected from a corpus of text, where the word appears many times. WEAT's effect size is then computed many times, with different embeddings each time, and a combined effect size is then calculated on it. As already mentioned by the original authors, even with only $2$ contextual embeddings collected per word in the WEAT stimuli, and each set of  $X, Y, A, B$  having only $5$ stimuli, $2^{5\cdot 4}$ possible combinations can be used to compute effect sizes.
    
    \item \textbf{Probe:} The entire example, or a specific word in the text, is probed for gender. A classifier is trained to learn the gender from a representation of the word or the text as extracted from a model. This can be done on examples where there is some gender labeling (for instance, the gender of the person discussed in a biography text) or when the text contains some target words, with gender context. Such target words could be ``nurse'' for female and ``doctor'' for male. Usually, the word probe refers to a classifier from the family of multilayer preceptron classifiers, linear classifiers included. The accuracy achieved by the probe is often used as a measure of how much gender information in embedded in the representations, but there are some weaknesses with using accuracy, such as memorization and other issues \cite{hewitt-liang-2019-designing, probing-belinkov}, and so MDL Probing is proposed as an alternative \cite{voita-titov-2020-information}, and the metric used is compression rate. Higher compression indicates more gender information in the representation.
    
    \item \textbf{Cluster:} It is possible to cluster the word embeddings or representations of the examples and perform an analysis using the gender labels just like in probing.
    
    \item \textbf{Nearest Neighbors:} As with probing, the examples and word representations can be classified using a nearest neighbor model, or an analysis can be done using nearest neighbors of word embeddings as done by \citet{gonen-goldberg-2019-lipstick-pig}.
    
    \item \textbf{Gender Space:} in the static embeddings regime, \citet{bolukbasi2016man} proposed to identify gender bias in word representations by computing the direction between representations of male and female word pairs such as ``he'' and ``she''. They then computed PCA to find the gender direction. \citet{basta2021extensive} extended the idea to contextual embeddings by using multiple representations for each word, by sampling sentences that contain these words from a large corpus. \citet{zhao-etal-2019-gender} performed the same technique on a different dataset. They then observed the percentage of variance explained in the first principal component, and this measure plays as a bias metric. The principal components can then be further used for a visual qualitative analysis by projecting word embeddings on the component space.
    
    \item \textbf{Cos:} in static word embeddings \cite{bolukbasi2016man}, this was computed as the mean cosine similarity between neutral words which might have some stereotype such as ``doctor'' or ``nurse'', and the gender space. \citet{basta2021extensive} computed it on profession words using extracted embeddings from a large corpus.
\end{enumerate}

\section{Statistical Fairness Metrics}
\label{app:statistical_metrics}

This section describes statistical metrics that are representative of many other fairness metrics that have been proposed in the field. \textit{separation} and \textit{sufficiency} fall under the definition of ``extrinsic performance'', specifically ``gap (Gender)'' while \textit{independence} falls under the definition of ``extrinsic prediction'', specifically ``pred gap''. Various numbers are generated by these metrics that describe differences between two distributions as measured by Kullbeck-Liebr divergence. We sum all the numbers to quantify bias in a single number.

Let $R$ be a model's prediction, $G$ the protected attribute of gender, and $Y$ the golden labels.

\paragraph{Independence} requires that the model's predictions are independent of the gender. Formally:

$P(R=r | Z=F) = P(R=r | Z=M)$

It is measured by the distributional difference between $P(R=r)$ and $P(R=r|Z=z)$ $\forall z\in \{M,F\}$.

\paragraph{Separation} requires that the model's predictions are independent of the gender \textit{given the label}. Formally:

$P(R=r | Y=y, G=F) = (R=r | Y=y, G=M) \forall y \in \mathcal{Y} $

It is measured by the distributional difference between $P(R=r | Y=y, Z=z)$ and $P(R=r | Y=y) \forall y \in \mathcal{Y}, \forall z \in \{M,F\}$ 

\paragraph{Sufficiency} requires that the distribution of the gold labels is independent of the model's predictions \textit{given the gender}. Formally:

$P(Y=y | R=r, G=F) = P(Y=y | R=r, G=M)$

It is measured by the distributional difference between $P(Y=y | R=r, Z=z)$ and $P(Y=y | R=r) \forall y \in \mathcal{Y}, \forall z \in \{M,F\}$ 

\section{Bias in Bios experiments}
\subsection{Implementation details}
\label{app:impl}

In this section we describe the metrics that were measured in the experiments on Bias in Bios, following \citet{our_paper}.

\paragraph{Performance gap metrics.} The standard measure for this task \cite{biasinbios} is the True Positive Rate (TPR) gap between male and female examples, for each profession $p$: $$TPR_p=TPR_{p_F}-TPR_{p_M}$$ and then compute the Pearson correlation between each $TPR_p$ and the percentage of females in the training set with the profession $p$. The result is a single number in the range of 0 to 1, with a higher value indicating greater bias. We measure the Pearson correlations of $TPR_p$, as well as of the False Positive Rate (FPR) and the Precision gaps. In addition, we sum all the gaps in the profession set $P$, thereby quantifying the absolute bias and not only the correlations, for example, for the TPR gaps: $\sum_{p\in P} TPR_p$.

\paragraph{Statistical fairness metrics.} We also measured three statistical metrics \cite{barocas-hardt-narayanan}, relating to several bias concepts: Separation, Sufficiency and Independence. A greater value means more bias. Detailed information on these metrics can be found in Appendix \ref{app:statistical_metrics}.

\subsection{Correlations between extrinsic and intrinsic metrics when measured on different datasets}
\label{app:correlations}

\begin{table*}

\centering

\begin{tabular}{ccc}
\toprule
\multicolumn{1}{l}{Metric} & \multicolumn{1}{l}{Bias in Bios (Original)} & \multicolumn{1}{l}{Winobias} \\
\midrule
TPR gap (P) & 0.304 & 0.022 \\
TPR gap (S) & 0.449 & 0.002 \\
FPR gap (P) & 0.120 & 0.030 \\
FPR gap (S) & 0.046 & 0.008 \\
Precision gap (P) & 0.063 & 0.013 \\
Precision gap (S) & 0.291 & 0 \\
Independence gap (S) & 0.382 & 0.005 \\
Separation gap (S) & 0.165 & 0.001 \\
Sufficiency gap (S) & 0.567 & 0.025 \\
\bottomrule
\end{tabular}

\caption{Correlations between intrinsic and extrinsic metrics. Original correlations are from \citet{our_paper}, our correlations are calculated with the intrinsic metric as measured on Winobias.}
\label{tbl:other_data_correlations}
\end{table*}

Table \ref{tbl:other_data_correlations} present the full results of our correlation tests, when intrinsic metrics was measured on a different dataset (Winobias) than the extrinsic metric (Bias in Bios). For all metrics, there is no correlation when we measured the intrinsic metric with a different dataset, although many of the metrics did correlate with the intrinsic metrics when measured on the same dataset as is originally done in \citeauthor{our_paper}.

\subsection{Statistics of the Dataset Before Balancing}
\label{app:bias-in-bios-stats}

Table \ref{tbl:bias-in-bios-stats} presents how the professions in Bias in Bios dataset \cite{biasinbios} are distributed, per gender. The gender was induced by the pronouns used to describe the person in the biography, thus it is likely the self-identified gender of the person described in it.
\begin{table*}[h]
\centering
\begin{tabular}{lrr}
\toprule
{} &     Females &     Males \\
\midrule
professor         & 45.10\% & 54.90\% \\
accountant        & 36.73\% & 63.27\% \\
journalist        & 49.51\% & 50.49\% \\
architect         & 23.66\% & 76.34\% \\
photographer      & 35.72\% & 64.28\% \\
psychologist      & 62.07\% & 37.93\% \\
teacher           & 60.24\% & 39.76\% \\
nurse             & 90.84\% &  9.16\% \\
attorney          & 38.29\% & 61.71\% \\
software\_engineer & 15.80\% & 84.20\% \\
painter           & 45.74\% & 54.26\% \\
physician         & 49.37\% & 50.63\% \\
chiropractor      & 26.31\% & 73.69\% \\
personal\_trainer  & 45.56\% & 54.44\% \\
surgeon           & 14.82\% & 85.18\% \\
filmmaker         & 32.94\% & 67.06\% \\
dietitian         & 92.84\% &  7.16\% \\
dentist           & 35.28\% & 64.72\% \\
dj                & 14.18\% & 85.82\% \\
model             & 82.74\% & 17.26\% \\
composer          & 16.37\% & 83.63\% \\
poet              & 49.05\% & 50.95\% \\
comedian          & 21.14\% & 78.86\% \\
yoga\_teacher      & 84.51\% & 15.49\% \\
interior\_designer & 80.77\% & 19.23\% \\
pastor            & 24.03\% & 75.97\% \\
rapper            &  9.69\% & 90.31\% \\
paralegal         & 84.88\% & 15.12\% \\
\bottomrule
\end{tabular}
\caption{Statistics of professions and genders as they appear in the Bias in Bios dataset.}
\label{tbl:bias-in-bios-stats}
\end{table*}

\subsection{Analysis of Precision (Pearson) Metric Behavior}
\label{app:precision-analysis}

This section discusses the results on the precision (Pearson) metric. This metric is measured as the Pearson correlation between the (i) percentage of women in the training set per profession (ii) precision gap between female examples and male examples in this profession (computed as $\text{precision}^F_{P} - \text{precision}^M_{P}$ per profession $P$).

\begin{figure}[H]
\begin{subfigure}[t]{\columnwidth}
  \includegraphics[width=\columnwidth]{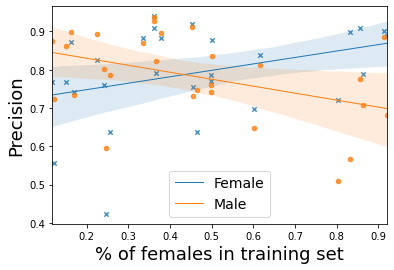}
  \caption{on an imbalanced dataset (same distribution as the training set).}
  \label{fig:precision_original}
\end{subfigure}

\begin{subfigure}[t]{\columnwidth}
  \includegraphics[width=\columnwidth]{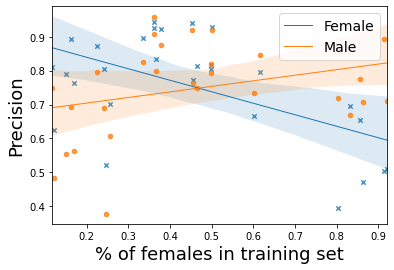}
  \caption{On a balanced (subsampled) dataset.}
  \label{fig:precision_balanced}
\end{subfigure}
  
\caption{Precision of female and male examples over all professions, as a function of the percentage of females in the training set.}
  \label{fig:precision}
\end{figure}

We observed a strong negative value for the metric when testing the model on a balanced test-set, but a positive value when testing on a test-set with the same distribution as the training-set (which is unbalanced). On an imbalanced test-set, the distribution is similar to the training-set and thus, for female examples, the precision is higher in a profession that is biased in favor of females (meaning, as the percentage of females in the training-set is higher). The same logic applies to the precision of males in professions skewed toward men. Therefore, the correlation is positive.  Conversely, if the test-set is balanced, the model's bias hurts the model's precision: as the profession is more biased, for instance the profession of "nurse", for which more than 90\% of the training examples are female, the model tends to assign females to the profession more often than necessary, hurting the precision. Figure \ref{fig:precision} demonstrate this point: on the imbalanced dataset (Plot \ref{fig:precision_original}), precision for females grows as the percentage of females is larger and precision for males reduces as the percentage of females is larger, and for a balanced dataset (Plot \ref{fig:precision_balanced}) the trend is reversed. For these reasons, we get a positive correlation with the precision gap on an imbalanced dataset and a negative correlation on a balanced dataset.

% \begin{figure}[h]
%   \includegraphics[width=.8\columnwidth]{precision_original_test_set.png}
%   \caption{on an imbalanced dataset (same distribution as the training set).}
%   \label{fig:precision_original}
% \end{figure}

% \begin{figure}[h]
%   \includegraphics[width=.8\columnwidth]{precision_subsampled_test_set.png}
%   \caption{On a balanced (subsampled) dataset.}
%   \label{fig:precision_balanced}
% \end{figure}

% We observed a negative correlation between the precision gap and the percentage of women in the training set when testing the model on a balanced dataset, and a positive correlation when testing the model on a biased dataset which has the same distribution as the training set. For the sake of the explanation, we will focus on a single profession with gender imbalance on the training set, for instance the profession ``nurse'', for which more than 90\% of the training examples are female.

% Precision is computed as $\frac{TP}{TP + FP}$. We compute precision per gender on the nurse examples. The gap

% In models biased toward a specific gender, for example nurse and female (since over 90\% of the nurse instances in the dataset are female), more examples of that gender are classified as that profession. This bias increases both TP and FP. In a balanced dataset, there are less positive examples (female nurses) than in a non-balanced dataset

% If a balanced dataset is used, the FP will be increased more, thereby decreasing precision and causing a negative correlation, whereas if an imbalanced dataset is used, the TP will be increased more, thereby increasing precision which results in a positive correlation.

\section{Full List of Reviewed Papers for Extrinsic Metrics Measurements}

Table \ref{tbl:extrinsic_count_full} presents the papers we reviewed and the amount of extrinsic metrics measured by them.

\begin{table*}

\centering

\begin{tabular}{lr}
\toprule
Paper & \multicolumn{1}{l}{Maximum \# of extrinsic} \\
& \multicolumn{1}{l}{metrics per task} \\
\midrule
\makecell[l]{\citet{bolukbasi2016man, zhang2018mitigating} \\ \citet{bordia-bowman-2019-identifying, ethayarajh-etal-2019-understanding} \\ \citet{sahlgren2019gender, karve-etal-2019-conceptor} \\ \citet{hall-maudslay-etal-2019-name, sedoc-ungar-2019-role} \\ \citet{kaneko-bollegala-2019-gender, liang-etal-2020-monolingual} \\ \citet{dev2020measuring, shin-etal-2020-neutralizing} \\ \citet{kaneko-bollegala-2021-dictionary}} & 0 \\
\midrule
\makecell[l]{\citet{zhao-etal-2017-men, zhao-etal-2018-gender} \\ \citet{li-etal-2018-towards, elazar-goldberg-2018-adversarial} \\ \citet{zmigrod-etal-2019-counterfactual, zhao-etal-2019-gender} \\ \citet{kumar-etal-2020-nurse, bartl-etal-2020-unmasking} \\ \citet{sen-etal-2021-counterfactually}} & 1 \\
\midrule
\makecell[l]{\citet{prost-etal-2019-debiasing, qian-etal-2019-reducing} \\ \citet{emami-etal-2019-knowref, habash-etal-2019-automatic} \\ \citet{dinan-etal-2020-queens, costa-jussa-de-jorge-2020-fine} \\ \citet{basta-etal-2020-towards}} & 2 \\
\midrule
\makecell[l]{\citet{park-etal-2018-reducing, stafanovics-etal-2020-mitigating} \\ \citet{saunders-byrne-2020-reducing, saunders-etal-2020-neural} \\ \citet{kaneko-bollegala-2021-debiasing, jin-etal-2021-transferability}} & 3 \\
\midrule
\citet{de-vassimon-manela-etal-2021-stereotype} & 6 \\
\midrule
\citet{our_paper} & 10 \\
\bottomrule
\end{tabular}

\caption{Papers about gender bias and the number of extrinsic metrics they measured per task. 0 means no extrinsic metrics were measured.}
\label{tbl:extrinsic_count_full}
\end{table*}

\end{document}